\def\year{2019}\relax
\documentclass[letterpaper]{article} 
\usepackage{aaai19}  
\usepackage{times}  
\usepackage{helvet}  
\usepackage{courier}  
\usepackage{url}  
\usepackage{graphicx}  
\usepackage{appendix}
\usepackage{amsmath,fullpage, graphicx, enumerate, verbatim}
\usepackage{amssymb}
\usepackage{algorithm}
\usepackage{dsfont}
\usepackage{algorithmic}
\usepackage{blindtext}
\title{Thompson Sampling for Pursuit-Evasion Problems}
\author{Zhen Li \\Department of Statistics\\ NC State University \And Nicholas J. Meyer\\Department of Statistics\\ NC State University \And Eric B. Laber\\Department of Statistics\\ NC State University \And Robert Brigantic\\Pacific
Northwest\\ National Laboratory}
\begin{document}
%

\maketitle
\begin{abstract}
Pursuit-evasion is a multi-agent sequential decision problem wherein a group of 
agents known as pursuers coordinate their traversal of a spatial domain 
to locate an agent trying to evade them.  Pursuit evasion problems arise
in a number of import application domains including defense and route planning.  
Learning to optimally coordinate pursuer behaviors so as to minimize time to capture
of the evader is challenging because of a large action space and sparse noisy state
information; consequently, previous approaches have relied primarily on heuristics.  
We propose a variant of Thompson Sampling for pursuit-evasion that allows for the application
of existing model-based planning algorithms.  This approach is general in that it allows
for an arbitrary number of pursuers, a general spatial domain, and the integration of 
auxiliary information provided by informants.  In a suite of simulation 
experiments, Thompson Sampling for pursuit evasion significantly reduces time-to-capture
relative to competing algorithms. 
\end{abstract}

\section{Introduction}
\label{sec:intro}

The general setup for pursuit evasion problems wherein multiple
agents coordinate their efforts to locate a shrewd
adversary is a useful model for important real-world
decision problems arising in security, law enforcement, and
wildlife management 
\cite{nahin2012chases}, \cite{fang2015security}.
Indeed, our interest in this problem is motivated by
our involvement with the adaptive search for nuclear materials in collaboration with the consortium for non-proliferation
enabling capabilities (CNEC). 

There is an extensive literature on pursuit-evasion
problems both from a theoretical and computational aspect
(see \cite{rodin1987pursuit}, \cite{yavin2014pursuit}
for an extensive list of references). Most closely
related to the problem we consider is the partially
observable Markov pursuit game studied by 
Hespanha et al. \shortcite{hespanha2000probabilistic}.
Heuristic search strategies in this domain are based on 
approximations of the belief state of the 
evader's location given current information;
e.g., the local-max heuristic
uses a one-step greedy optimization of the
probability of capture in the next time step whereas
the global-max heuristic moves pursuers toward
the posterior mode o the evader's location
\cite{hespanha2000probabilistic},  \cite{vidal2002probabilistic}.
Kwak \& Kim \shortcite{kwak2014policy} used a weighted combination of local- and global-max with the weights tuned using 
reinforcement learning.

We present a variant of Thompson Sampling for pursuit-evasion
that comprises the following steps at each time point: 
(i) computing
a posterior distribution over the space of possible
evader strategies; (ii) truncating the tails from this
posterior and then sampling a strategy from the
resultant truncated distribution; and (iii) using model-based
planning to estimate the optimal pursuer strategy.  
The truncation in (ii) can be seen as a mechanism
to limit exploration to make the method `safe,' i.e.,
to avoid potentially catastrophic action selection
\cite{garcia2015comprehensive}.  
The proposed algorithm performs favorably relative
to competitors in terms of time-to-capture
in a suite of simulation experiments including
pursuit-evasion over a grid and the coordination 
of ghost behavior in the class arcade game
Pac-Man.

\section{Setup and Notation}
\label{sec:setup}
We consider a pursuit-evasion problem evolving in discrete time 
over a spatial domain represented (or approximated) by a fixed network.  
We consider a team of $K$ pursuers coordinating their movements across the network to locate a 
single evader that is moving to avoid the pursuers and reach a 
goal node in the network.  The initial location of the evader and goal node
are unnecessarily known to the pursuers in our formulation.  If the pursuers intercept the evader before the
evader reaches a goal node, they receive
a positive reward signal which may depend on time to capture or other attributes
of their search path, e.g., paths taken etc.  At each time point, the pursuers
inspect nearby nodes (they have a limited radius of vision) for the evader 
and may also receive information about the evader's location from
benevolent informant; this latter feature is designed to reflect intelligence 
in a defense application.

Let \(\mathcal{L} = \lbrace 1, \ldots, L \rbrace\) be the set of
nodes in the network and let \(\Omega \in \lbrace 0, 1 \rbrace^{L \times L}\) 
be its adjacency matrix, i.e.,
\(\Omega_{\ell, \ell'} = 1\) if locations \(\ell\) and \(\ell'\) are
connected ($\ell$ is also connected to itself).  At each time point
\(t \in \mathcal{T} = \lbrace 0, 1, 2, \ldots\rbrace\), each agent 
selects a node from among the neighbors of their current location 
to which to move; a node is defined as its own neighbor to allow an agent
to remain in the same location for multiple time steps.  Let
\(E_t \in \mathcal{L}\) denote the location of the evader at time \(t\)
and \(G \subset \mathcal{L}\) the set of the evader's goal locations.  Thus, 
if \(E_t \in G\) before the evader is captured the game ends and the evader is declared
the winner.  Let
\(W_t =  \left(W_t^1, \ldots, W_t^K\right) \in \mathcal{L}^K\) denote the locations of
the $K$ pursuers at time $t$. Let $d(\ell, \ell)$ denote the graph distance between
nodes $\ell$ and $\ell'$.  If the event $\min_{k}d(W_k^t, E_t) \le 1$ occurs before
the event $E_t \in G$ then the evader is said to be caught, the game ends, and the pursuers are declared
the winners.  
Define $C_t$ to be an indicator that the game has not ended at time $t$;
thus, the duration of the game is $T = \min\left\lbrace t\,:\,C_t = 1\right\rbrace$.  
Let $Y_t$ denote a momentary reward for the pursuers, e.g.,  a small negative constant while the game is ongoing and a large negative constant
if the evader reaches its goal.  
In addition,
at each time $t$, the pursuers may receive information from an informant in the form of
a region $D_t \subseteq \mathcal{L}$ known to contain the evader at time $t$; for simplicity,
we assume that informant information is completely reliable though this can be relaxed.  
For notational convenience, when no informant information is provided we code $D_t \equiv 
\mathcal{L}$. Also, we denote $R_t$ as the event that the pursuers obtain the informant region $D_t$ at time $t$.  
The information available to the pursuers at time $t$ is therefore
\(H_t=\{R_0,W_0, C_0, Y_0,\ldots, Y_{t-1}, R_t,W_t, C_t\}\). 
Let $S_t = \left\lbrace H_t, E_0,\ldots, E_t\right\rbrace$ denote
the complete state of the system at time $t$.  

At each time point, the pursuers and evader select a neighboring node
to move to; however, the proposed methodology can be extended to
a richer set of actions, e.g., in the context of tracking nuclear material, the
set of actions might include planting a 
remote sensor.  Let $\mathcal{B}_{\mathcal{L}}$ denote the set of
distributions over $\mathcal{L}$.
A strategy, $\pi_E$, is an infinite sequence of 
functions $\pi_{E,t}:\mathrm{dom}\,S_t\rightarrow \mathcal{B}_{\mathcal{L}}$ such
that $\pi_{E,t}(S_t)$ has 
support only on the neighbors of $E_t$. Let $\Pi_E$ denote the set of allowable 
evader strategies.  Similarly, let 
$\mathcal{B}_{\mathcal{L}^K}$ denote the space of distributions over
$\mathcal{L}^K$ and define 
a strategy,
$\pi_{W}$, for the pursuers to be a sequence of maps $\pi_{W,t}:\mathrm{dom}\,H_t
\rightarrow \mathcal{B}_{\mathcal{L}^K}$ such that $\pi_{W,t}(H_t)$ has support only on the neighbors of $W_t$.  Let $\Pi_{W}$ denote the set of allowable pursuer strategies.  
In some applications, the evader may only have access to a coarse 
summary of $S_t$, as we shall see, such constraints can be imposed through the class of candidates strategies considered for the evader.  

For each $\pi_E\in\Pi_E$ and $\pi_W\in\Pi_W$, we define $V(\pi_E, \pi_W) =
\mathbb{E}^{\pi_E,\pi_W}\left(\sum_{t=0}^{T}\gamma^{t-1}Y_t\right)$, where
$\mathbb{E}^{\pi_E,\pi_W}$ denotes expectation with respect to the distribution
induced by the strategies $(\pi_E, \pi_W)$ and $\gamma\in (0,1]$ is a 
discount factor.  
Given a strategy for the evader, $\pi_{E}\in\Pi_E$, an optimal pursuer strategy 
$\pi_{W}^*\in\Pi_{W}$ satisfies $V(\pi_E, \pi_W^*) \ge V(\pi_E, \pi_W)$
for all $\pi_W\in\Pi_W$.

\section{Estimating the Evader's Strategy}
\label{sec:posterior}
If the evader's strategy is known, the pursuers can use standard methods from reinforcement
learning to construct an estimator of $\pi_W^{*}$ \cite{sutton1998introduction}\cite{powell2007approximate} \cite{si2004handbook} \cite{busoniu2010reinforcement} \cite{szepesvari2010algorithms}. 
Unfortunately, in practice the evader's strategy is not generally known.  One approach
is to approximate a Nash-equilibrium \cite{littman1994markov}\cite{hu2003nash}\cite{wang2003reinforcement}.
However, while such game-theoretic solution concepts are appealing in some contexts, such
equilibria need not be unique and furthermore fail to exploit non-equilibria behavior 
\cite{gmytrasiewicz2005framework}; this latter point is particularly relevant in 
pursuit-evasion problems arising in defense and security applications in which evaders are likely following pre-defined strategies, employing heuristics, or acting erratically.  
We instead propose to model the evader's strategy using accumulating data and then to apply
a variant of Thompson Sampling wherein, at each time $t$, 
an evader's strategy along with any other requisite system
dynamics are sampled from a posterior given $H_t$ and then these dynamics are used
in model-based planning algorithm to estimate an optimal pursuer strategy 
\cite{gopalan2015thompson}.  

Let $\Pi_E$ be a pre-specified class of candidate policies for the evader and let
$\rho(\pi_E)$ denote a prior distribution over this class. 
The class $\Pi_{E}$ could
be finite, finite-dimensional, or even infinite-dimensional though in most applications this class
is heavily informed by domain expertise and finite-dimensional.  
At each time
$t$, the posterior distribution $p(\pi_E|H_t)$ is used to quantify uncertainty about the evader's
strategy given the history available to the pursuers; thus, $\pi_E$ is treated as a parameter 
indexing the model whereas $\pi_W$ is under control of the pursuers.
Under mild regularity conditions, 
we provide closed-form expressions for $p(\pi_E|H_t)$ and the posterior distribution of the evader's location; these expressions
are of interest in their own right as estimators of such probabilities are used in heuristic search strategies \cite{hespanha2000probabilistic}.   

Let $\mathbf{p}_{t,\pi_W}(H_{\tau}, \pi_E)$ be a $|\mathcal{L}|$-dimensional vector where its $r$th component equals $P_{\pi_W}(E_t=r|H_\tau,\pi_E)$, the probability the evader's location is $r$ at time $t$ given the history $H_\tau$ under policies $\pi_E$ and $\pi_W$. $T_t(H_t,\pi_E)$ is the transition matrix for the evader's strategy $\pi_E$ at time $t$ where $(T_t(H_t,\pi_E))_{\ell',\ell}=P(E_{t+1}=\ell'|E_{t}=\ell,H_t, \pi_E)$, which is given for $\forall\; \pi_E\in \Pi_E$; let $F$ be the standardization operator on the $|\mathcal{L}|$-dimensional
positive orthant, i.e., $x \mapsto x/\sum_{j}x_j$. At each time $t$ define  
\begin{equation}
D_{t,\pi_E}=D_t\cap (\underset{\ell\in W_t}{\cup} \mathcal{A}_\ell)^c \cap G_{\pi_E}^c,
\end{equation}
where $G_{\pi_E}$ is the evader's goal set under strategy $\pi_E$. Let  \(I_{D_{t,\pi_E}}\) denote an $|\mathcal{L}|\times |\mathcal{L}|$ diagonal matrix such that
the $\ell$th diagonal element is \(1\) if $\ell\in D_{t,\pi_E}$ and \(0\) otherwise.

\newtheorem{assumption}{\bf Assumption}[section]
\begin{assumption}
\label{A1}
For \(D_t\subset \mathcal{L}\), $P_{\pi_W}(R_t|W_t,E_t=r,Y_{t-1},H_{t-1},\pi_E)$ is a constant for $\forall \;r\in D_t$ and $\forall \;\pi_E \in \Pi_E$.
\end{assumption}

 Intuitively, this assumption states that the probability that the informant provides
 non-trivial information about the location of the evader does not depend on the
 evader's location or their strategy; 
 this assumption can be relaxed to handle the setting where there
 are multiple informants more prone to report in different regions of the network.  
\newtheorem{lemma}{\bf Lemma}[section]
\begin{lemma}
For $\forall \;\pi_E \in \Pi_E$ and $\forall \;\pi_W\in \Pi_W$,
$$
\begin{aligned}
&\mathbf{p}_{t,\pi_W}(H_{t-1},\pi_E)\\
=&T_t(H_{t-1},\pi_E) F(I_{D_{t-1,\pi_E}} \mathbf{p}_{t-1,\pi_W}(H_{t-2},\pi_E)).
\end{aligned}
$$
Moreover, $\mathbf{p}_{t,\pi_W}(H_{t-1},\pi_E)$ is constant for $\forall \; \pi_W\in \Pi_W$.
\end{lemma}

\newtheorem{corollary}{\bf Corollary}[section]
\begin{corollary}
For $\forall \;\pi_E \in \Pi_E$ and $\forall \;\pi_W \in \Pi_W$, $$\mathbf{p}_{t,\pi_W}(H_t,\pi_E)=F(I_{D_{t,\pi_E}} \mathbf{p}_{t,\pi_W}(H_{t-1},\pi_E)).$$ Moreover, $\mathbf{p}_{t,\pi_W}(H_t,\pi_E)$ is constant for $\forall \; \pi_W\in \Pi_W$.
\end{corollary}
Through the recursion in the preceding lemma, one can derive  $\mathbf{p}_{t,\pi_W}(H_{t-1},\pi_E)$ 
from the probability vector of the evader's initial location, $\mathbf{p}_{0}$. 
\newtheorem{thm}{\bf Theorem}[section]
\begin{thm}
Let $\pi_E$ have prior $\rho(\pi_E)$, then $\forall \; \pi_W \in \Pi_W$,
$$
\begin{aligned}
p_{\pi_W}(\pi_E|H_t)
\propto \prod_{i=0}^t P_{\pi_W}(E_i\in D_{i,\pi_E}|H_{i-1}, \pi_E) \rho(\pi_E).
\end{aligned}
$$ 
\end{thm}

 Note that $P_{\pi_W}(E_i\in D_{i,\pi_E}|H_{i-1}, \pi_E)$ can be derived from $\mathbf{p}_{i,\pi_W}(H_{i-1},\pi_E)$ from which $p_{\pi_W}(\pi_E|H_t)$ can be obtained. We can see that $\mathbf{p}_{t,\pi_W}(H_{t-1},\pi_E)$, $\mathbf{p}_{t,\pi_W}(H_t,\pi_E)$ and $p_{\pi_W}(\pi_E|H_t)$ do not depend on $\pi_W$, so we suppress $\pi_W$ in the notation for simplicity. 
 Moreover, it means no matter which search strategy the pursuers follow, we can use the preceding relationships to derive the posterior of the evader's location and strategy at each time $t$.

\section{Estimating the Optimal Search Strategy}
\label{sec:optimal-search}

Suppose that the optimal search strategy for the pursuers, say $\pi_W^*$, were known, 
the optimal action for the pursuers at time $t$ would thus be 
\begin{equation}
\label{arg}
\underset{a}{\text{argmax}} \;\mathbb{E}^{\pi_W^*,\pi_E} \left[ \sum_{v \geq 0} \gamma^{v} Y_{t+v} \bigg| H_t, A_t=a \right],
\end{equation}
where $A_t$ is the pursuers' action at time $t$. 
Furthermore, it can be seen that the expectation in (\ref{arg}) is equal to 
\begin{equation}
\label{Qori}
\begin{aligned}
&\sum_{r\in \mathcal{L}} \mathbb{E}^{\pi_W^*,\pi_E} \left[ \sum_{v \geq 0} \gamma^{v} Y_{t+v} \bigg|E_t=r, H_t, A_t=a \right]\\
\times&P_{\pi_W^*}(E_t=r|H_t,A_t=a,\pi_E).
\end{aligned}
\end{equation}
Under strategies $\pi_E$ and $\pi_W^*$, $P_{\pi_W^*}(E_t=r|H_t,A_t=a)$ equals $\lbrace \mathbf{p}_{t}(H_t,\pi_E)\rbrace_r$ which is obtained in the previous section since the action $A_t=a$ is chosen deterministically. Thus, we only need to compute the Q-function 

\begin{equation}
\label{Q}
\begin{aligned}
&Q^{\pi_W^*,\pi_E}_t(r,H_t,a)\\
=&\mathbb{E}^{\pi_W^*,\pi_E} \Bigg[ \sum_{v \geq 0} \gamma^{v} Y_{t+v} \bigg|E_t=r, H_t, A_t=a \Bigg]\\
=&\mathbb{E}^{\pi_W^*,\pi_E} \Bigg[ \sum_{v = 0}^{n-1} \gamma^{v} Y_{t+v} + \sum_{v \geq n} \gamma^{v} Y_{t+v} \bigg|\\
&E_t=r, H_t, A_t=a \Bigg].
\end{aligned}
\end{equation}

Evaluating the Q-function is computationally burdensome and grows
exponentially in the number of time points.  Truncating the evaluation
at \(n\) points can make the computation manageable.  However, this
will result in a loss of precision.  An alternative is to evaluate the
Q-function using an \(n\)-point expansion and for all further points
approximate using a heuristic strategy.  The expensive part of the
evaluation is calculating the max at each point because it requires
recursive enumeration of all possible paths.  However, with a
heuristic strategy, we only need to enumerate all possible paths for $n$ steps and follow the heuristic strategy for the other steps, which can reduce the computational burden to a great extent. Thus, we approximate
(\ref{Q}) by 
\begin{equation}
\label{Qhat}
\begin{aligned}
&\widehat{Q}^{\pi_W^*,\pi_E}_t(r,H_t,a)\\
=&\mathbb{E}^{\pi_W^*,n,\pi,\pi_E} \Bigg[ \sum_{v = 0}^{n-1} \gamma^{v} Y_{t+v} + \gamma^n \sum_{v \geq 0}  \gamma^{v} Y_{t+n+v} \bigg|\\ 
&E_t=r, H_t, A_t=a \Bigg],
\end{aligned}
\end{equation}
where $\mathbb{E}^{\pi_W^*,n,\pi, \pi_E}$ is the expectation if pursuers follows the optimal strategy $\pi_W^*$ before time $t+n$ and a heuristic strategy $\pi$ after that. 

If $n=1$, (\ref{Qhat}) is one-step look-ahead and the method to obtain the optimal action is known as the rollout algorithm with the rollout policy $\pi$ \cite{sutton1998introduction}. If $n\geq 2$, we can apply the heuristic search method to compute $(\ref{Qhat})$ when additional assumptions are made.  If $\gamma^n \sum_{v \geq 0}  \gamma^{v} Y_{t+n+v}$ is
\(o_p(1)\), then the approximation error can be made arbitrarily small
by increasing \(n\).  Thus the choice of the heuristic strategy has
little impact for large \(n\).

To approximate the optimal strategy, we use a heuristic strategy that
moves pursuers in the direction of the locations with the largest
posterior coverage.  Let \(\pi_H\) be the heuristic strategy and $\pi_E$ be the evader's strategy.  This
heuristic strategy is defined as
\begin{equation*}
  \pi_H(H_t) = \underset{\substack{a_1,\ldots,a_K \\
      a_i \in \mathcal{A}_{w_i}}}{\arg\min} \; \sum_{i = 1}^K
  \text{dist}(a_i, q_i(H_t))
\end{equation*}
where $\mathbf{q}(H_t)=(q_1(H_t),\ldots,q_K(H_t))$ are the positions of the posterior with largest cumulative
coverage at the next time step. The \(i\)th location, \(q_i(H_t)\), is
assigned to pursuer \(i\) by minimizing the cumulative distance
between each pursuer and their assigned target location. These
positions are defined as
\begin{equation*}
\begin{aligned}
  &\mathbf{q}(H_t) = \underset{\substack{\lbrace
      \ell_1,\ldots,\ell_K\rbrace \subset \mathcal{L}\\ \ell_k \ne \ell_{k'}}}{\arg\max} \; \Bigg\{
  \sum_{j = 1}^K \text{dist}(w_j, \ell_j)^{-1}  \\
  &\mathds{1} \left[
    \sum_{j = 1}^K (\mathbf{p}_{t+1,\pi_E})_{\ell_j} = \underset{\substack{\lbrace \widetilde{\ell}_1,
        \ldots, \widetilde{\ell}_K \rbrace
        \subset \mathcal{L} \\ \widetilde{\ell}_k \ne \widetilde{\ell}_{k'}}}{\max} \; \sum_{j = 1}^K
    (\mathbf{p}_{t+1}(H_t,\pi_E))_{\widetilde{\ell}_j}\right] \Bigg\}.
    \end{aligned}
\end{equation*}

Since the game model is complex and computing the joint probability of all events is not generally feasible; instead, we apply a simulation-based method to approximate (\ref{Qhat}). 
To compute (\ref{Qhat}) requires that the pursuers follow the optimal strategy $\pi_W^*$ from the beginning of the game to simulate a process which has the history $H_t$ and $E_t=r$ at time $t$.  This is not generally possible as $\pi_W^*$ is unknown and the probability of the event $\{E_t=r,H_t\}$ occuring is typically vanishingly small when $t$ is large. To solve this problem, we assume the evader's location $E_t$ and the pursuers' history $H_t$ provide a sufficient summary of the overall history $S_t$ so that the evader's strategy $\pi_E$ is determined only by $E_t$ and $H_t$. 
\begin{assumption}
\label{Markov}
$(Y_t,E_{t+1},R_{t+1},W_{t+1},C_{t+1},\ldots)$ given $S_t$ depends on $S_t$ only through $(E_t,H_t)$, for $t\in \mathcal{T}$.
\end{assumption}

\begin{assumption}
\label{ePolicy}
For $t\in \mathcal{T}$, $$\pi_{E,t}(S_t)=\pi_{E,t}(E_t,H_t).$$
\end{assumption}
Assumption \ref{Markov} is weaker than the Markov assumption as it does not consider the influence of the pursuers' history $H_t$ on the future. Assumption \ref{ePolicy} guarantees that the pursuers can simulate the evader's action at time $t$ given only $E_t=r$ and $H_t$. Moreover, the transition matrix $(T_t(H_t,\pi_E))_{\ell',\ell}=P(E_{t+1}=\ell'|E_{t}=\ell,H_t, \pi_E)$ is determined by $\pi_E$. These two assumptions hold in some common cases and we will discuss them further in Section 5. With these two assumptions, the pursuers can conduct simulation starting from time $t$ given $E_t=r$ and $H_t$ instead of from the beginning of the game. Moreover, (\ref{Qhat}) becomes 
\begin{equation}
\label{Qhatnew}
\begin{aligned}
&\widehat{Q}^{\pi_W^*,\pi_E}_t(r,H_t,a)
=\mathbb{E}^{\pi_W^*,\pi_E} \Bigg[ \sum_{v = 0}^{n-1} \gamma^{v} Y_{t+v} + \\
&\gamma^n Q_{t+n}^{\pi,\pi_E}(E_{t+n},H_{t+n},\pi(H_{t+n})) \bigg|E_t=r, H_t, A_t=a \Bigg].
\end{aligned}
\end{equation}

Then we can apply the heuristic search method to compute (\ref{Qhatnew}) where the values of leaf nodes are $Q_{t+n}^{\pi,\pi_E}(E_{t+n},H_{t+n},\pi(H_{t+n}))$ and they are backed up to the current state $(E_t=r,H_t)$ at the root. More details can be found in \cite{sutton1998introduction}.
In our experiment, we approximate (\ref{Qhatnew}) using a heuristic to reduce computation time. We consider all possible paths for pursuers from time $t$ to $t+n-1$ (taking action $a$ at $t$) and then let pursuers follow the heuristic strategy $\pi$. We run $s$ independent simulations for each path and obtain the average values of $\sum_{v\geq 0}\gamma^{v} Y_{t+v}$ for each path. The path with the largest average values implements the optimal strategy approximately and the largest average values is an approximation for (\ref{Qhatnew}). Note that when $n=1$, the method is actually the rollout algorithm. After we compute (\ref{Qhatnew}), the optimal action at time $t$ can be obtained by (\ref{arg}) and (\ref{Qori}).

The discussion above is to estimate the optimal strategy for the pursuers when the the evader's strategy is given. In practice, the the evader's strategy is not generally known but in 
Thompson Sampling, a candidate strategy for the evader can be sampled from a (possibly truncated) posterior over $\Pi_E$. 
The proposed algorithm, including the discussed heuristics, is given Algorithm \ref{thompson}.
Note that in lines 7 and 15 in the algorithm, we can apply any other strategy, e.g., one estimated using reinforcement learning, the local-max strategy, or the global-max strategy. 
In the next section, we examine the empirical performance of Algorithm \ref{thompson} 
through simulation of pursuit-evasion on a grid and a modification of the classic
arcade game Pac-Man.

\section{Numerical Results and Analysis}
\subsection{Random-Walk-with-Drift Evader}
In this case, the class of evader strategies
which we consider is a random walk with drift towards the goal states.
Each strategy in the class is indexed by two values: the evader's goal node,
\(G \in \mathcal{L}\), and the amount of drift, \(\xi \in [0, 1]\). 
With probability
\(\xi\), the evader takes the action that moves the evader closest to the goal
state (with ties broken uniformly at random); otherwise, 
the evader takes a random action with uniform probability. For simplicity, we assume that the pursuers have prior belief in a finite number of possible goals $PG \subset \mathcal{L}$ and possible drifts $PD \subset [0, 1]$.  Thus $\Pi_E$ is a finite set $\{ \pi_1, . . . , \pi_B \}$.  However, the true goal ($TG$) and the true drift ($TD$) of the evader may or may not belong to $PG$ and $PD$ respectively.

\begin{figure}
  \centering
  \includegraphics[width=0.3\textwidth]{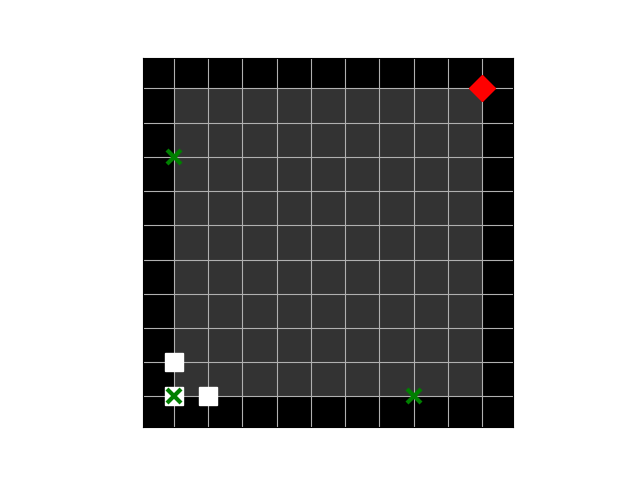}
  \caption{\label{fig:sim-setup}Schematic for starting positions of
    all units and evader's goal locations when $m=10$, $K=3$.  The red diamond is the
    starting evader's location $\{99\}$.  The white squares are the starting locations of the pursuers $\{0,1,10\}$.
    Green crosses are the possible locations of the evader's goal $\{0,7,70\}$.}
\end{figure}

The network is an
\(m\times m\) grid.  The pursuers and evader start in the opposite
corners of the grid and pursuers know the evader's initial location $D_0$. Figure \ref{fig:sim-setup} shows the starting positions of all units and the evader's goal locations when $m=10$, $K=3$. The informant region $D_t$ is governed by both the pursuers' vision with radius $v$ and information occasionally provided by an informant. Whether information is obtained from the informant at time $t$ is determined by the random variable
\(O_t = \mathds{1}_{\sum_{v = 1}^t X_v > t}\) where
\(X_v \overset{iid}{\sim} \text{exp}(\lambda)\), $\lambda=0.3$ in our experiment. If $O_t=1$, the pursuers learn the evader's location --- i.e., each of the locations in the quadrant where the evader is currently located is appended to the informant region. If $O_t=0$, no information is obtained from the informant at time $t$. For this episodic task, define $T$ to be the (random) time point at which the game ends (that is, at which either the evader is caught or reaches the goal) and set the discount factor $\gamma=1$. As for the outcome, $Y_t=-1$ for $t=0,1,\ldots, T-1$. If the evader is caught by the pursuers at $T$, $Y_t=-1$ and if the evader reaches goal node $G$ at $T$, $Y_T=-100$.  Thus in order to maximize cumulative reward, the pursuers must capture the evader as soon as possible and keep the evader from reaching his goal.

Finally, a brief remark on the reasonableness of assumptions (\ref{A1}), (\ref{Markov}), (\ref{ePolicy}). For Assumption (\ref{A1}), we have $P_{\pi_W}(R_t|W_t,E_t=r,Y_{t-1},H_{t-1},\pi_E)=1$ for $\forall\;r\in D_t$ and $\forall\;\pi_E\in \Pi_E$, no matter if $O_t$ is 0 or 1 (the proof is simple and we omit it due to space). For Assumption (\ref{Markov}), it is evident the future event after time $t$ is independent of the overall history $S_t$ given the pursuers' history $H_t$ and the evader's location $E_t$. Moreover, the evader's strategy only depends on his current location so Assumption (\ref{ePolicy}) holds.

As measures of performance, we consider the 
proportion of episodes in which the evader is 
captured ($C_1$), the average time at which the 
pursuer is captured ($T$), and the proportion of 
episodes in which the evader was captured via the shortest possible path given the evader's trajectory ($C_2$). 

The number of pursuers $K$ is set to be 2 and grid 
sizes $m=10,20$ are considered. For the following 
experiments, we specify a uniform prior over $\Pi_E$ and set the vision radius $v=2$ and $TD=0.75$. If $m=10$, we set $TG=7$ and if $m=20$, we set $TG=14$. We consider the 12 experiment settings shown in Table \ref{tab:summary}. Experiment A is a benchmark search strategy in which the pursuers know the exact location of the evader and select their actions to 
minimize the distance to the evader (with ties broken uniformly at random). In Experiment B, pursuers make decisions according to minimax Q-learning \cite{littman1994markov} and know the exact location and the goal of the evader. Experiments 1-10 use our proposed method, with varying values of $m$, $PG$, $PD$, and $n$ (and with $d = 0.9$ in each case).

\begin{figure}
  \centering
  \includegraphics[width=0.35\textwidth]{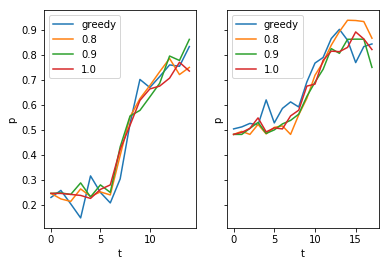}
  \caption{\label{fig:truncate_post}The proportion of the true evader's strategy sampled by Truncated Thompson sampling with different truncating coefficients ($d$) under $m=10$, $n=0$, $TG=7$, $TD=0.75$. The left side is for the setting $PG=\{7,70\}$, $PD=\{0.25, 0.75\}$, $v=2$ while the right side is for $PG=\{7,70\}$, $PD=\{0.75\}$, $v=0$. 500 games are simulated for each setting.}
\end{figure}

Figure \ref{fig:truncate_post} displays the proportion of times that the evader's strategy is sampled by Truncated Thompson sampling over the course of an episode (and demonstrates that this tends to increase to 1). Table \ref{tab:vision} shows that the pursuers are more likely to capture the evader and capture times are shorter as the vision radius $v$ increases. Performance in Experiments 1-10 is better on our measures of performance than in baseline Experiments A and B.
Experiments 1 and 6 show the superior performance of the proposed reinforcement learning method given the true strategy for the evader. (Note that even the true optimal search strategy may not catch the evader through the shortest path due to the randomness of the evader's movement.)  When multiple strategies are given prior probability but the class of possible strategies still contains the evader's strategy, performance degrades relative to Experiments 1 and 6 but it still superior to the baseline methods (Experiments 2-4 and 7-9).
Finally, Experiments 5 and 10 show two cases in which 
reasonable performance is attained even when the evader's strategy is not in the class of possible strategies assumed by the pursuers. 

\subsection{Pac-Man}

The game Pac-Man \cite{truemanhistory} can be seen as a pursuit-evasion problem. In the game, there are four ghosts and Pac-Man navigating a maze. The goal of the four ghosts is to catch Pac-Man, while Pac-Man's goal is to eat as many dots as possible without being captured by the ghosts.  If Pac-Man eats one of the four lage power dots near the corners of the maze, the ghosts are temporarily slowed and vulnerable to being eaten by Pac-Man.

\begin{figure}
  \centering
  \includegraphics[width=0.2\textwidth]{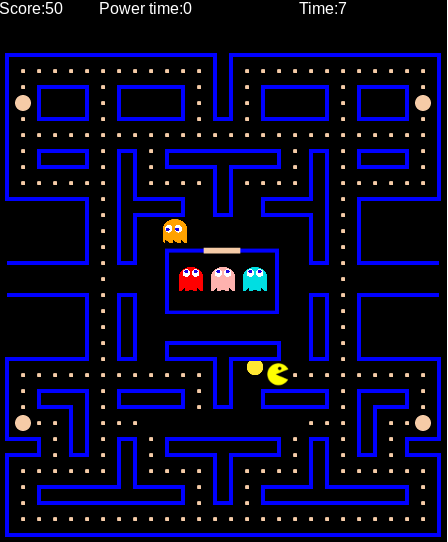}
  \caption{\label{fig:pacman}A screen shot for the game Pac-Man.}
\end{figure}

In order to further study our method,
we implemented a JavaScript version of Pac-Man in which the ghosts (pursuers) could follow one of several 
pursuit strategies (Figure \ref{fig:pacman} displays a screenshot of the game). $\Pi_E$ consists of three strategies we devised for Pac-Man. The first is a pure random walk. The second is defined such that when the distance between Pac-Man and one of the ghosts is less than some value $\delta$, the Pac-Man will move in the opposite direction of the closest ghost; otherwise Pac-Man will move uniformly at random. The third is the same as the second except that when the distance between Pac-Man and one of the ghosts is at least $\delta$, Pac-Man moves towards the closest Pac-Dot with probability $\xi$ and moves randomly otherwise. 
Thus, we write $\Pi_E=\{\pi_1,\pi_2,\pi_3^\xi\}$, where $\xi\in [0,1]$ is fixed. 

Both Pac-Man and the ghosts know the location of each dot and whether it has been eaten or not. Thus the pursuers' history $H_t$ should include the dots' history (i.e., the location of each dot and whether it has been eaten at each time up until $t$). Moreover, the Pac-Man knows the ghosts' locations exactly, while the ghosts only have vision radius $v$. The informant region $D_t$ for the ghosts is determined by their vision and the dots' history (as Pac-Man's exact location is made known to the ghosts when a dot is eaten). The outcome $Y_t$ is defined similarly as that in Section 5.1 except there is not a goal location for Pac-Man. Under these simulation settings, assumptions (\ref{A1}), (\ref{Markov}) and (\ref{ePolicy}) still hold (the proof of this is simple and we omit it here). In the experiment, the ghosts' prior on Pac-Man's 
strategy is uniform on 
$\Pi_E=\{\pi_1,\pi_2,\pi_3^{0.8}\}$, whereas the true 
strategy followed by Pac-Man is $\pi_3^{1.0}$; thus the set of possible strategies considered by the pursuers 
does not include the true strategy.

As before, We compare our method with a search strategy in which the ghosts know Pac-Man's exact location and select their actions to minimize the distance to Pac-Man. Scores for Pac-Man and capture time ($T$) are shown in Table \ref{tab:pac}. The results show that our method performs slightly better than the benchmark strategy even if the ghosts give zero prior weight to Pac-Man's true strategy. However, the strategy considered is the true strategy for Pac-Man with some randomness and it explains why our method performs well. (In practice, including in $\Pi_E$ strategies which include randomness may improve robustness, as this may account for e.g. accidental missteps by the evader.)

\section{Discussion}
\label{sec:conclustion}
We formalized the problem of pursuit and evasion
and developed a general framework for rigorously constructing and testing
estimators of the evader's strategy and the pursuers optimal search strategy.  We
demonstrated methods for estimating the evader's strategy and the optimal search strategy using
a rollout-based approximation to the Q-function. The proposed estimators performed well across
a variety of settings though in many settings greedy application of Thompson Sampling (i.e.,
using the posterior mode for $\pi_E$) performs well. 
Because the proposed framework is Bayesian, it can seemlessly incorporate prior 
knowledge which may be especially beneficial in defense applications where the 
behavior of the adversary has been studied extensively.  

There are a number of interesting directions for future work. 
One direction is to add complexities to more closely mimick real-life
search problems.  The abilities of all search units could be expanded by
providing new actions in addition to movement. For example, the
pursuers might have a ``scan'' action which forces them to remain still,
but allows them to capture the evader if they are \(k\) edges away
instead of just \(1\).  Another example is a ``dash'' action which
allows a pursuer to move at a faster rate, but the ability to
detect the evader is less reliable. 
Another area of future research is prioritization of capture zones.
In real world applications, capturing an evader could have negative side
effects.  For example, someone carrying nuclear materials might
detonate on capture.  Thus, it is important to try and capture the
evader in an area that will minimize damage.  Incorporating this
prioritization into the pursuer search strategy could lead to interesting
new methodologies.

\subsubsection*{Acknowledgements}
We would like to thank the Consortium for Nonproliferation Enabling
Capabilities for funding this work.  We would also like to thank the
Pacific Northwest National Laboratory for providing expert knowledge
on the pursuit and evasion problem.

\begin{algorithm}[H]
	\begin{algorithmic}[1]
	\STATE From domain knowledge, the pursuers are given a class of strategies $\Pi_E$ wherein lies the evader's strategy possibly. Also, each element $\pi_E \in \Pi_E$ has a prior $\rho(\pi_E)$.
	\STATE Observe $H_0=(R_0,W_0,C_0)$.
	\STATE Initialize the prior probability vector of $E_0$, $\mathbf{p}_0 \in \mathbb{R}^{|\mathcal{L}|}$.
	\STATE Initialize $D_{0,\pi_E}=D_0\cap (\underset{l\in W_0}{\cup} \mathcal{A}_l)^c \cap G_{\pi_E}^c$.
	\STATE Initialize $K(H_0|\pi_E)=P(E_0\in D_{0,\pi_E})$, for $\pi_E\in \Pi_E$.
	\STATE Sample $\tilde{\pi}_E\sim \rho(\pi_E)$.
	\STATE Initialize $n$ and $\mathbf{p}_0(H_0,\tilde{\pi}_E)=F(I_{D_{0,\tilde{\pi}_E}}\mathbf{p}_0)$. Obtain $r=\text{argmax}_{r\in \mathcal{L}} (\mathbf{p}_0(H_0,\tilde{\pi}_E))_r$. Compute (\ref{Qhatnew}) with simulation for $\forall \;a$ and select $A_0=a$ with the largest values in (\ref{Qhatnew}) as the optimal action for pursuers.
		\FOR {iteration $t=1,2,\ldots$}
		\STATE Observe $(R_t,W_t,C_t)$.
		\STATE $\mathbf{p}_t(H_{t-1},\pi_E)=T_t(H_{t-1},\pi_E)
		\mathbf{p}_{t-1}(H_{t-1},\pi_E)$
		for $\pi_E\in \Pi_E$.
		\STATE $D_{t,\pi_E}=D_t\cap (\underset{l\in W_t}{\cup} \mathcal{A}_l)^c \cap G_{\pi_E}^c,$ for $\pi_E\in \Pi_E$.
		\STATE
		$$
		K(H_t|\pi_E)=\sum_{r\in D_{t,\pi_E}} (\mathbf{p}_{t}(H_{t-1},\pi_E))_r K(H_{t-1}|\pi_E),
		$$
		for $\pi_E\in \Pi_E.$
		\STATE
		$
		p(\pi_E|H_t)
		=\frac{K(H_t|\pi_E)\rho(\pi_E)}{\int_{\pi_E'\in \pi_E} K(H_t|\pi_E')p(\pi_E')},\;\text{for}\; \pi_E\in \Pi_E.
		$
		\STATE Sample $\tilde{\pi}_E\sim p(\pi_E|H_t)$.
		\STATE $\mathbf{p}_t (H_t,\tilde{\pi}_E)=F(I_{D_{t,\tilde{\pi}_E}}\mathbf{p}_{t}(H_{t-1},\tilde{\pi}_E))$ and $r=\text{argmax}_{r\in\mathcal{L}}(\mathbf{p}_t (H_t,\tilde{\pi}_E))_r$. Compute (\ref{Qhatnew}) with simulation for $\forall \;a$ and select $A_t=a$ with the largest values in (\ref{Qhatnew}) as the optimal action for pursuers.
		\ENDFOR
	\end{algorithmic}
	\caption{Thompson sampling for the pursuit-evasion problem.}
	\label{thompson}
\end{algorithm}
\begin{table}[]
		\centering
		\begin{tabular}{c|c|c}
			\hline
			$v$ & 1 & 2   \\
			\hline
			
			$C_1$ &0.71 & 0.90\\
			\hline
			$T$ &11.76 &10.84\\
			\hline
		\end{tabular}
	    \caption{Performance of the pursuers with different vision radius when $m=10$, $PG=\{7,70\}$, $PD=\{0.75\}$, $TG=7$, $TD=0.75$ and $n=1$. 100 games are simulated.}
		\label{tab:vision}
	\end{table}
\begin{table}[]
		\centering
		\begin{tabular}{c|c|c|c|c}
			\hline
			Exp. & $m$ & $PG$ & $PD$ & $n$   \\
			\hline
			A & 10 &$*$ & $*$ & $*$\\
			\hline
			B & 10& $\{7\}$ & $*$ & $*$\\
			\hline
			1 & 10 &$\{7\}$ & $\{.75\}$ & 2 \\
			\hline
			2 & 10& $\{7,70\}$ & $\{.25, .75\}$ & 0 \\
			\hline
			3 & 10& $\{7,70\}$ & $\{.25, .75\}$ & 1 \\
			\hline
			4 & 10& $\{7,70\}$ & $\{.25, .75\}$ & 2 \\
			\hline
			5 & 10& $\{8,80\}$ & $\{.2, .5\}$ & 2 \\
			\hline
			6 & 20& $\{14\}$ & $\{.75\}$ & 2 \\
			\hline
			7 & 20& $\{14,140\}$ & $\{.25, .75\}$ & 0 \\
			\hline
			8 & 20& $\{14,140\}$ & $\{.25, .75\}$ & 1 \\
			\hline
			9 & 20& $\{14,140\}$ & $\{.25, .75\}$ & 2 \\
			\hline
			10 & 20& $\{12,120\}$ & $\{.5, .9\}$ & 2 \\
			\hline
			\hline
			Exp. & $C_1$ & $T$ & $C_2$ &  \\
			\hline
			A &  .72 & 11.08 (0.17)& $*$ &\\
			\hline
			B & .43 & 12.28 (0.26)& $*$\\
			\hline
			1 & 1& 10.24 (0.11) & .95&\\
			\hline
			2 &  .81 & 11.21 (0.19) & .57 \\
			\hline
			3 & .89 & 10.71 (0.17)& .78 \\
			\hline
			4 &  .95 & 10.66 (0.14) & .84 \\
			\hline
			5 &  .89 & 10.73 (0.17) & .78 \\
			\hline
			6 & 1 & 21.60 (0.25) & .88\\
			\hline
			7 & .84 & 24.30 (0.44)& .46 \\
			\hline
			8 & .90 & 23.22 (0.38) & .52 \\
			\hline
			9 & .92 & 23.21 (0.48) & .58 \\
			\hline
			10 & .84 & 24.23 (0.47)& .50 \\
			\hline
		\end{tabular}
	    \caption{A summary of the simulation results for different settings when the truncation coefficient is $0.9$. Experiment A and B represent the benchmark and the minimax Q-learning strategies respectively. $n=0$ represents the heuristic strategy defined in Section 4. For each setting, 100 and 50 games are simulated for $m=10$ and $m=20$ respectively. Standard errors are shown in the brackets.}
		\label{tab:summary}
	\end{table}
\begin{table}[]
		\centering
		\begin{tabular}{c|c|c}
			\hline
			 & $T$ & Score   \\
			 \hline
			 TTS & 127 (8) & 874 (57)\\
			 \hline
			 Benchmark & 131 (10) & 900 (59)\\
			\hline
		\end{tabular}
	    \caption{The capture time and scores for the Truncated Thompson Sampling method ($d=0.9$) and the benchmark method in the Pac-Man game. For each setting, 50 games are simulated. Standard errors are shown in the brackets.}
		\label{tab:pac}
	\end{table}

\clearpage
\bibliographystyle{aaai}
\bibliography{mybib}

\end{document}